\documentclass[10pt,twocolumn,letterpaper]{article}

\usepackage{wacv}
\usepackage{times}
\usepackage{epsfig}
\usepackage{graphicx}
\usepackage{amsmath}
\usepackage{amssymb}
\usepackage{algorithm}
\usepackage{algorithmic}
\usepackage{multirow}

\DeclareMathOperator*{\argmin}{argmin}

\def\wacvPaperID{525} 

\wacvfinalcopy 

\ifwacvfinal
\fi

\ifwacvfinal
\usepackage[breaklinks=true,bookmarks=false]{hyperref}
\else
\usepackage[pagebackref=true,breaklinks=true,colorlinks,bookmarks=false]{hyperref}
\fi

\ifwacvfinal
\else
\fi

\begin{document}

\title{Deep Poisoning: Towards Robust Image Data Sharing against Visual Disclosure}

\author{
Hao Guo$^1$,
Brian Dolhansky$^2$,
Eric Hsin$^2$,
Phong Dinh$^2$,
Cristian Canton Ferrer$^2$,
Song Wang$^1$\\
$^1$ University of South Carolina,
$^2$ Facebook AI\\
\vspace*{1mm}
}

\maketitle

\begin{abstract}
    Due to respectively limited training data, different entities addressing the same vision task based on certain sensitive images may not train a robust deep network. This paper introduces a new vision task where various entities share task-specific image data to enlarge each other's training data volume without visually disclosing sensitive contents (e.g. illegal images). Then, we present a new structure-based training regime to enable different entities learn task-specific and reconstruction-proof image representations for image data sharing. Specifically, each entity learns a private Deep Poisoning Module (DPM) and insert it to a pre-trained deep network, which is designed to perform the specific vision task. The DPM deliberately poisons convolutional image features to prevent image reconstructions, while ensuring that the altered image data is functionally equivalent to the non-poisoned data for the specific vision task. Given this equivalence, the poisoned features shared from one entity could be used by another entity for further model refinement. Experimental results on image classification prove the efficacy of the proposed method.

\end{abstract}

\section{Introduction}
Deep networks, e.g. Convolutional Neural Networks (CNNs), have achieved state-of-the-art results on many computer vision tasks~\cite{he2017mask,he2016deep,krizhevsky2012imagenet,liu2016ssd,long2015fully,redmon2016you,ren2015faster}, which can be used in many critical production systems~\cite{chen2015deepdriving,geiger2012we,liuLQWTcvpr16DeepFashion}. Traditionally, training of these networks requires large-scale task-specific datasets with many images~\cite{ILSVRC15}. However, for certain vision tasks with restricted images, one entity (institution/company) may not properly collect sufficient images for robust deep model learning. To deal with this, various entities could enlarge training data volume by sharing image data with each other.
Nevertheless, the task-specific images may contain overly sensitive visual contents that should not be spread, making the sharing of raw images inappropriate. Thus, this paper introduces a new task on vision integrity of preventing image data sharing from visually disclosing sensitive image contents.

The introduced task of image data sharing against visual disclosure includes two objectives: 1) the image data (e.g. convolutional features) shared by different entities should be same task-specific, so that 
one entity could utilize data shared from various entities for the model learning;
2) one entity should not be able to visually observe the image contents by recovering images from the shared data,
e.g. image reconstruction -- defined as visual disclosure of sensitive image contents.
There are broad practical applications of this task. For instance, 
deep model based child exploitation image (CEI) detection and terrorist propaganda image detection help comfort the online social community. The model learning requires a large volume of training images, which may not be easily collected by an individual company, due to the highly sensitivity of the images. The introduced task allows various companies to share image data for sensitive vision tasks without spreading uncomfortable images, which may also violate laws.

\begin{figure*}[!t]
\begin{center}
\includegraphics[width=0.9\textwidth]{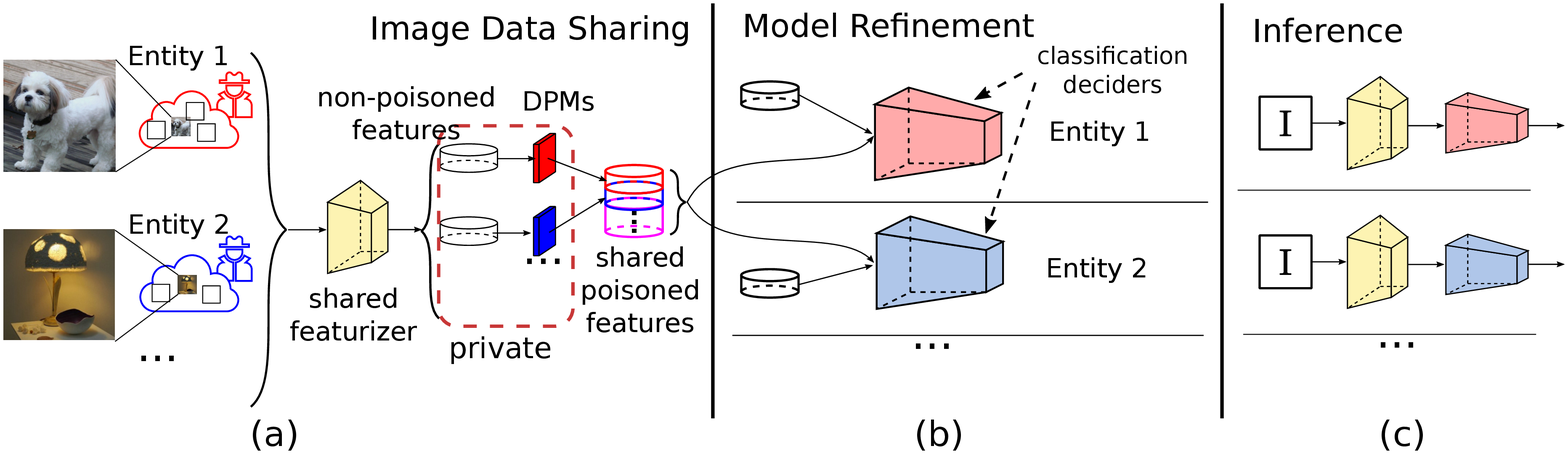}
\end{center}
\caption{
An illustration of image data sharing for collaboratively addressing the same task. (a) Each entity uses an individual private DPM for feature poisonining -- training details discussed later; (b) One entity uses its own data and image data shared by others to refine respective models, e.g. classification decider; (c) Using the refined models for image inference. 
}
\label{fig:fig_intro}
\end{figure*}

Different from the task of privacy preserving, which disentangles utility information from privacy information (e.g. facial expression v.s. face identity) on a constant dataset, the introduced task focuses on a collaborative dataset, allowing collaborators (entities) securely sharing and using the image data to refine deep model learning without undesired image spreading. Privacy-preserving methods usually convert the raw images to image representations by a certain operation, such as anonymization~\cite{li2019anonymousnet, sun2018hybrid}, encryption~\cite{gilad2016cryptonets, kim2018secure, yonetani2017privacy}, privacy-preserving representation learning~\cite{roy2019mitigating, creager2019flexibly, xiao2019adversarial, pittaluga2019learning, chen2018vgan}.
For the introduced task, shared image data should be in the same feature space (the same task specific), which requires the above operation being shared among collaborators. Thus, one entity may reverse the operation, e.g. like a black box, to reconstruct the original images from data shared by other entities, leading to visual disclosure of image contents.
Besides, comparing with federated learning~\cite{mcmahan2017communication} using extra hardware for simultaneous model learning from multiple entities without sharing image data, the introduced task enables image data sharing without visual disclosure and allows various entities to train models individually, leading to enlarged applications.



\newpage
\par This paper focuses on the regime of preventing visual disclosure of sensitive image contents (e.g. private faces, illegal images) during image data sharing  for a particular vision task. To illustrate, we take the image classification as an example of the vision task to be addressed -- various entities collaborate on training models on image classification by sharing their respectively collected image data to each other. First, one entity pre-trains a specific image classification network, and split the network at a specific point into the image featurizer (consisting of certain starting layers) and the classification decider (the remaining layers), similar to~\cite{osia2017privacy, osia2018deep}. Entities can use the shared featurizer to produce feature maps for their images. Instead of sharing these deep features, which can be easily reconstructed to the original images by reversing the featurizer, we design a regime that requires each entity to learn a respective Deep Poisoning Module (DPM) with various architectures to poison the deep features. By means of adversarial training, each DPM is optimized to ensure that the poisoned features are functionally equivalent to the original features for image classification, while they can not be recovered to the original images by reversing the featurizer. As shown in Fig.~\ref{fig:fig_intro}, with keeping the DPM in private -- a partial-release strategy, the poisoned features shared from one entity can be used by another entity for refining the classification decider. Meanwhile, without accessing to the DPM, others can not reconstruct images from the poisoned features, which ensures the image data sharing against visual disclosure.

Finally, we conduct experiments to verify that the proposed DPM can prevent visual disclosure of sensitive contents during the image data sharing with a minimal loss in image classification performance. By simulating the process of collaborators exploiting the shared image data from other entities, our experiments demonstrate that the proposed framework is an effective way for image data sharing for collaboration without visually disclosing sensitive contents.



\section{Related Work}
\label{sec:related_work}

To protect information privacy, privacy-preserving data publishing (PPDP)~\cite{fung2010privacy,xu2014a} has been studied for a long time. It collects a set of individual records and publishes the records for further data mining, without disclosing individual attributes such as gender, disease, or salary~\cite{agrawal2000privacy,lindell2000privacy,mendes2017privacy,pinkas2002cryptographic,vaidya2002privacy}.
Existing work on PPDP mainly focuses on anonymization~\cite{chen2013privacy,grau2016logical,wong2006alpha} and data slicing~\cite{li2010slicing}. While it usually handles individual records related to identification, it is not explicitly designed for general high-dimensional data, such as images.

Recently, the privacy issue has been attracting an increasing attention by the computer vision and deep learning community. It is an important task for vision ethics. For example, MS-Celeb-1M~\cite{guo2016ms,pearson_2019} and Duke MTMC~\cite{ristani2016MTMC} were withdrawn from public release due to privacy issues. Existing methods on preserving privacy in images and videos usually alter the images or learn image representations so that the private information is degraded in the data. 
Intuitive perturbations, such as blurring and blocking~\cite{li2017blur,mahendran2015understanding,mcpherson2016defeating}, modify the images to reduce privacy information.
De-identification methods~\cite{li2019anonymousnet, sun2018hybrid} partially alter images, for example by obfuscating faces. 
Encryption-based approaches~\cite{gilad2016cryptonets, kim2018secure, yonetani2017privacy} train models directly on encrypted data.
Optimal transformations for producing super low-resolution images or videos in order to avoid leaking sensitive information are learned in ~\cite{ryoo2017privacy}.
Inspired by Generative Adversarial Nets (GAN)~\cite{goodfellow2014generative}, adversarial approaches~\cite{kim2019training, li2019deepobfuscator, oh2017adversarial, raval2017protecting, ren2018learning, speciale2019privacy, wang2019privacy, wu2018towards,roy2019mitigating, creager2019flexibly, xiao2019adversarial, pittaluga2019learning, chen2018vgan} learn deep obfuscators for images or corresponding convolutional features.

Even though the above methods achieve promising results on preserving privacy, they can not address the introduced task of image data sharing against visual disclosure. 
To enable that one entity utilize image data shared from various entities, they should apply the same privacy-preserving operation to convert images to the same feature space. However, the information to be shared in the image data and the visual depictions to be protected may not be disentangled efficiently as privacy-preserving methods distinguish utility and privacy. Thus, the image data shared by one entity could be reconstructed to visual depictions (the original images) by reversing the shared privacy-preserving operation (e.g. by means of reversing a black box), which still leads to undesired image spreading. 


The more similar task to the introduced one is the federated learning~\cite{mcmahan2017communication} -- multiple entities simultaneously update a model without sharing data to each other. 
Different from the introduced task, it usually learns a model on a third-party hardware, e.g. an extra server, or from cloud side. Image data are not required to be shared between entities (clients). This well addresses the data island issue, but limits the independence of each entity. Compared with federated learning, this paper proposes a new solution for various entities collaborating with each other to address the same vision task based on image data sharing with not visually disclosing sensitive image contents.

\section{Proposed Regime}

\subsection{Overview}

The overall image data sharing regime, proposed in this paper, for collaborations between various entities consists of three steps: 
\begin{itemize}
    \item[1] An initial deep network is pre-trained by one entity for a specific vision task, from which a featurizer is extracted to convert images to convolutional features;
    \item[2] A private Deep Poisoning Module is learned by each entity to poison the image features for image data sharing with not visually disclosing sensitive contents;
    \item[3] Each entity exploits image data shared from others for deep model refinement to better address the specific vision task.
\end{itemize}

\subsection{Step 1: Initial Pre-training}
\label{sec:step1}

\begin{figure}
    \centering
    \includegraphics[width=0.88\linewidth]{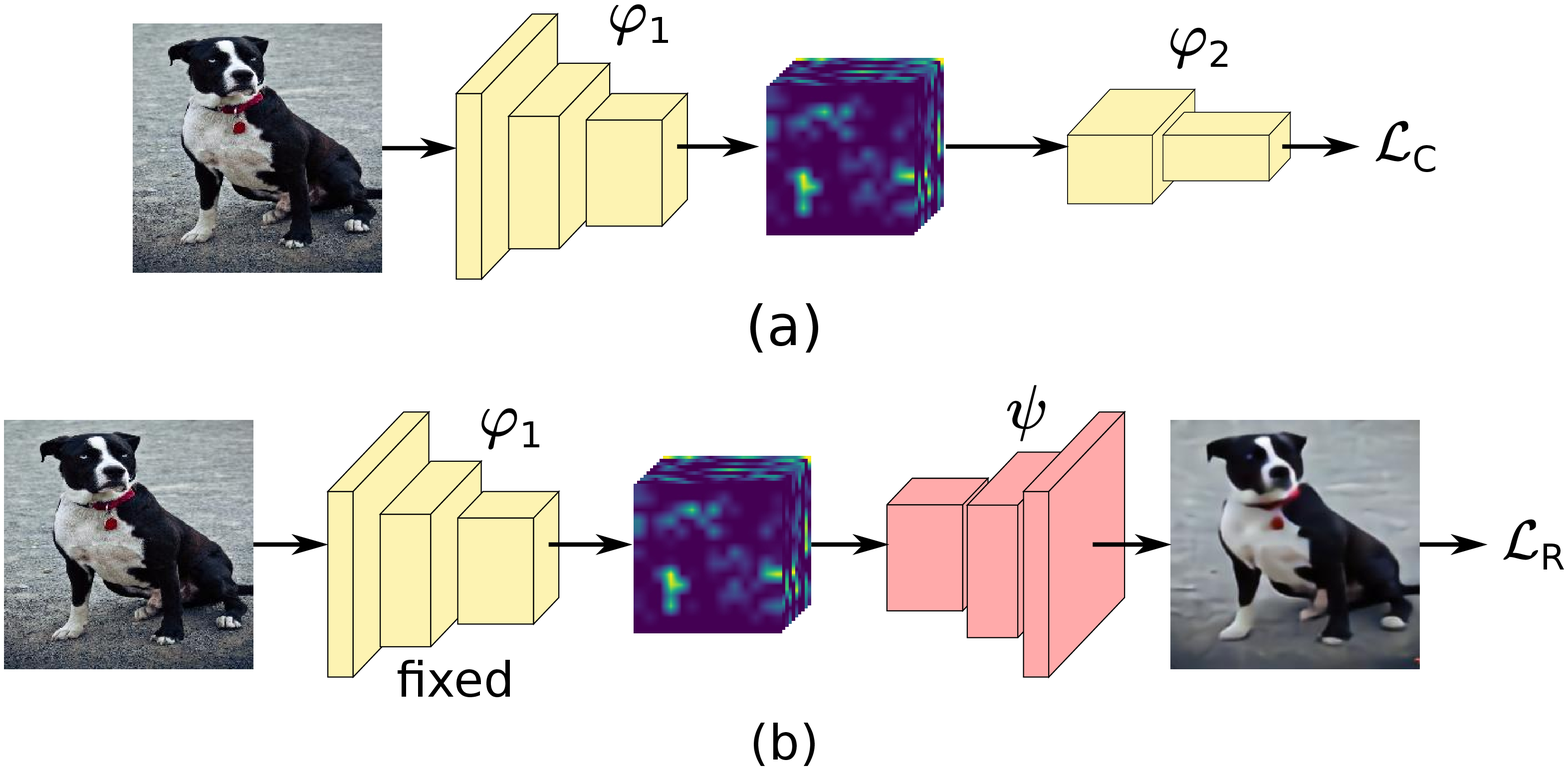}
    \caption{Illustration of the initial pre-training.}
    \label{fig:fig_init_train}
\end{figure}

The collaboration between entities aims to expand the volume of training images for each entity addressing the specific vision task, i.e. image classification. To achieve this goal, we expect that all entities share the same operation to converting raw images to certain image representations. At the beginning of the collaboration, one of the entities with the most of training samples pre-trains a classification network $\Phi$ based on a specific architecture, such as VGGNet~\cite{simonyan2014very}, ResNet~\cite{he2016deep}, ResNeXt~\cite{xie2017aggregated} or DenseNet~\cite{huang2017densely}, and the conventional classification loss is adopted for model optimization:
\begin{equation}
    \label{eq_cls_loss}
    \mathcal{L}_C(x, y_i)=-\log{\left (e^{p(\Phi(x)=y_i)}  / \sum_{j}{e^{p(\Phi(x) = y_j)}} \right )
    } ,
\end{equation}
where $y_i$ represents the annotation of the input image $x$. As illustrated in Fig~\ref{fig:fig_init_train}(a), the pre-trained model $\Phi$ is divided into two sequential modules: the image featurizer $\varphi_1$ and the classification decider $\varphi_2$:
\begin{equation}
    \label{eq_cls_split}
    \Phi(x) = \varphi_2(\varphi_1(x)).
\end{equation}
Based on the same featurizer, various entities can convert their raw images to the image data in the same feature space. However, the convolutional feature maps can be easily reconstructed to the original images, due to the rich visual information remembered in the features. As shown in Fig.~\ref{fig:fig_init_train}(b), the image reconstructor $\psi$ is learned to reverse the featurizer by minimizing the L1 loss between the original image and the reconstructed image:
\begin{equation}
    \label{eq_l1_loss}
    \mathcal{L}_R = \left \| x - \psi(\varphi_1(x)) \right \|_1.
\end{equation}

After the pre-training, each entity share the image featurizer, classification decider and the image reconstructor with fixed parameters.

\subsection{Step 2: DPM Training}
\label{sec:step2}

\begin{figure*}
    \centering
    \includegraphics[width=0.65\linewidth]{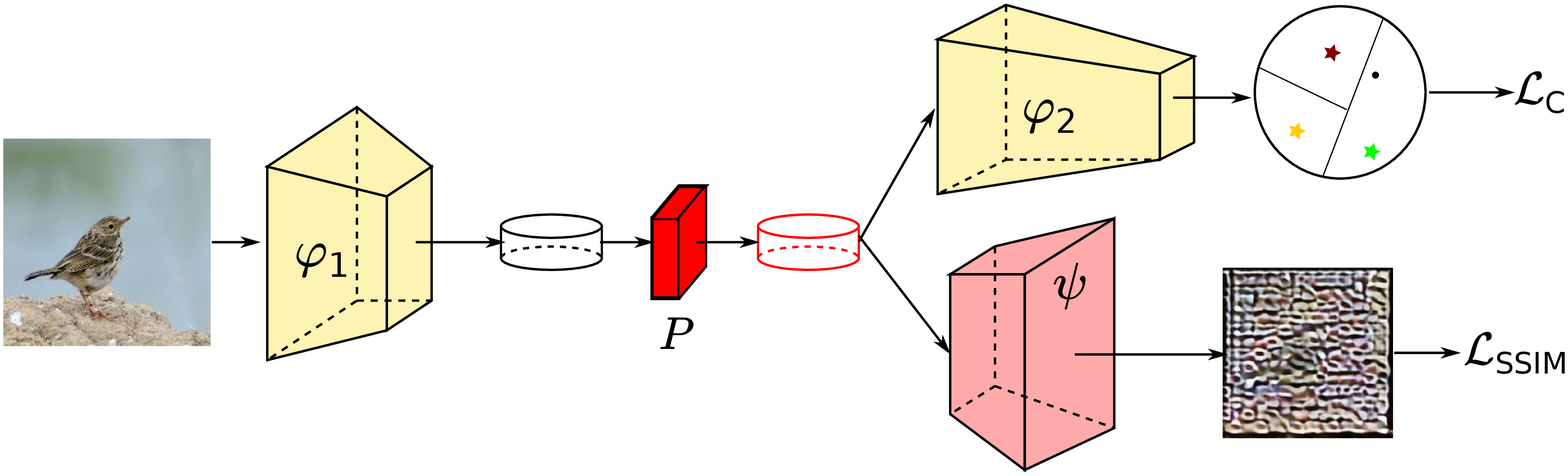}
    \caption{The learning framework of a DPM. The parameters of $\varphi_1$, $\varphi_2$ and  $\psi$ are fixed during DPM training. }
    \label{fig:fig_pf_framework}
\end{figure*}

The reconstruction in Fig.~\ref{fig:fig_init_train}(b) shows that if one entity shares the conventional features directly to collaborators, the collaborators can easily recover the original images, leading to visually spreading images. To deal with this issue, this paper proposes the regime that each entity designs a private Deep Poisoning Module (DPM) to poison the image features before sharing. 
The DPM consists of a sequential of convolutional layers and activation layers. It is an extra opertaion that disturbs the original features.
Each entity could keep its DPM in private, so that other entities are denied to learn an image reconstructor to reverse the poisoned features. The private DPM makes the ground truth (the pairs of poisoned features and original images) unavailable for reconstructor learning. Meanwhile, each DPM is also learned to ensure that the poisoned features are functionally equivalent to the original features (produced directly from the image featurizer) for the classification decider.

Specifically, as shown in Fig.~\ref{fig:fig_pf_framework}, the learning of a DPM is achieved by adversarially optimizing two loss functions. By fixing the parameters of $\varphi_1$ and $\varphi_2$ and minimizing the classification loss in Eq.~(\ref{eq_cls_loss}), the poisoned features achieve \textit{classification equivalence} to the original features:
\begin{equation}
    \label{eq_cls_approx}
    \varphi_2(P(\varphi_1(x))) = \varphi_2(\varphi_1(x)), 
\end{equation}
where $P$ is the private DPM.

To make the poisoned features not be reversed to the original images by the pre-trained image reconstructor $\psi$ (an inverse of the shared featurizer $\varphi_1$), the DPM is also learned to achieve reconstruction disparity -- it makes the reconstructed image from poisoned features $\psi(P(\varphi_1(x))) $ visually dissimilar to the original one $x$, by minimizing the Structural Similarity Index Measure (SSIM)~\cite{hore2010image, wang2004image}. We use ${SSIM}( \cdot, \cdot) \in [0, 1]$ between two images as the loss function:
\begin{equation}
    \label{eq_rec_loss}
    \mathcal{L}_{SSIM} = {SSIM}(\psi(P(\varphi_1(x))), x).
\end{equation}
Besides, the parameters of the pre-trained image reconstructor is also fixed during DPM learning.

Then, the DPM is finally optimized by the linear combination of the classification loss and the reconstruction loss:
\begin{equation}
    \label{eq_optim}
    \theta_P = \argmin_{\theta_P} \mathcal{L}_C + \lambda \argmin_{\theta_P} \mathcal{L}_{SSIM},
\end{equation}
where $\lambda$ is a hyper-parameter to balance two losses, and $\theta_P$ represents the parameters of the DPM.

Generally, with fixed parameters in the pre-trained featurizer, classification decider and image reconstructor, DPMs of various entities allows various collaborators to convert their images to deep representations in the same feature space (i.e. classification equivalence) by feeding them to partially different operations -- the same featurizer but different DPMs, respectively. This classification equivalence ensures that the poisoned features from different entities can be combined for classification decider refinement, while the partial sharing of the operations (keeping DPMs in private) makes the inverse of the operation infeasible, which defend against the image reconstruction from the poisoned features. Compared with deep obfuscated representations~\cite{li2019deepobfuscator, oh2017adversarial, raval2017protecting, ren2018learning, speciale2019privacy, wang2019privacy, wu2018towards}, which retain image classification-related information and suppress the reconstruction-related information adversarially, the proposed private DPM framework relies on a specific structure to deny the image reconstructor learning and thus is more reliable for preventing visual disclosure. 
To retain the functional ability of convolutional features for image classification, 
Since the reconstruction-related information can not be eliminated thoroughly to retain the functional ability of convolutional features for image classification, once the obfuscator is shared to collaborators for making classification equivalence, they can still reverse the image features to original images.

\subsection{Step 3: Classification Decider Refinement}
\label{sec:step3}

With the shared image featurizer $\varphi_1$ and private DPMs $\{P_1, P_2, ...\}$, each entity featurizes and poisons its images, and share the poisoned features to other collaborators, as shown in Fig.~\ref{fig:fig_intro}(a). Then, each collaborator could combine the shared image features (poisoned) and its own image features (original) to refine the classification decider, as illustrated in Fig.~\ref{fig:fig_intro}(b). The sequential combination of the shared featurizer and its refined classification decider allows an entity to form a more robust deep model for image classification task in Fig.~\ref{fig:fig_intro}(c).









\section{Experiments}

In this section, we first conduct experiments to prove that the proposed DPM can effectively poison the image features for image data sharing as expected: 1) the poisoned features are functionally equivalent to the original ones for a specific vision tasks, i.e. image classification; 2) the poisoned image features can not be reconstructed to the original images for visual disclosure of sensitive image contents. Then, we compare the the proposed DPM with existing methods, including conventional perturbations and adversarial obfuscation, to clarify its reliability. Furthermore, by  simulating entities exploiting poisoned and shared image features from others to refine model learning, we also verify the effectiveness of using DPM for image data sharing among collaborators addressing the same vision task.

\subsection{Configurations}
Instead of collecting some specific sensitive images, e.g. CEIs, which may be inappropriate for exhibition in conference papers, we use the widely-used ImageNet dataset~\cite{ILSVRC15} (1000-category classification) for simulation. Specifically, to simulate the task of various entities collaborating to address the image classification task, the dataset is randomly split into two sets, each with images of exclusive 500 categories (around 640K training images). One of the 500-category image sets $\mathbb{S}$ simulates the image datasets for addressing the image classification task, while the other one $\mathbb{Q}$ simulates a public image dataset. We suppose that image contents in $\mathbb{S}$ should not be visually disclosed during the image data sharing. 
Both $\mathbb{S}$ and $\mathbb{Q}$ contain training and validation subsets.

Due to its general applicability for computer vision tasks, we adopt a ResNet~\cite{he2016deep} architecture as the backbone network.
Following the expressions in Table 1 of~\cite{he2016deep}, we use $conv[\cdot]\_[\cdot]$ to represent the hook point that splits the architecture into the image featurizer and the classification decider.
For example, $conv4\_1$ indicates that the featurizer consists of the layers from the start of the architecture until the first building block of $layer4$ in the ResNet architecture.

\subsection{DPM for Image Data Sharing against Visual Disclosure}

\subsubsection{Effectiveness of DPM}


Suppose we are an individual entity with the collected image set $\mathbb{S}$ and would like to share these image data to others for image classification model training, i.e. the classification decider in Fig.~\ref{fig:fig_intro}. Initially, following the Step 1 in Sec.~\ref{sec:step1}, we conventionally train the 500-category image classification models based on ResNet50 and ResNet101, respectively. Given the input images with spatial dimension $224\times 224$, the top-1 and top-5 precision on the validation set of $\mathbb{S}$ achieved by ResNet50 are 79.39\% and 94.18\%, respectively, while that achieved by ResNet101 are 81.13\% and 95.03\%, respectively.

Then, the layer point $conv4\_1$~\cite{he2016deep} for both models are selected for network split. The image reconstructor $\psi$ to reverse the featurizer $\varphi_1$ is composed from the inverse of bottleneck blocks in ResNet~\cite{he2016deep}. Specifically, two inverse bottleneck blocks (CONV$1\times1$ -- BN -- CONV$3\times3$ -- BN -- CONV$1\times1$ -- ReLU) are used before upscaling the spatial dimension of feature maps by the factor of 2. As the feature maps with spatial dimension $14\times14$ are upscaled to $224\times 224$, a CONV$1\times1$ -- BN -- ReLU -- CONV$1\times1$ module is appended to produce the reconstructed image. This image reconstructor is learned as shown in Fig.~\ref{fig:fig_init_train}(b) and optimized by the loss in Eq.~(\ref{eq_l1_loss}). Since we would like to simulate the process of an entity reversing the featurizer instead of the specific set of image features, the image reconstructor is trained on the image set $\mathbb{Q}$.

\begin{table}[!t]
\begin{center}
\begin{tabular}{l|l|c|c}
\hline
\multirow{2}{*}{Backbone} & \multirow{2}{*}{Acc. Metric (\%)} & \multicolumn{2}{l}{Convolutional Features} \\ \cline{3-4} 
                  &                   & Original  & Poisoned  \\ \hline
\multirow{2}{*}{ResNet50} & top-1    &  79.39    & 78.88     \\ \cline{2-4} 
                  &         top-5    &  94.18    & 94.11     \\ \hline
\multirow{2}{*}{ResNet101}& top-1    &  81.13    & 80.78     \\ \cline{2-4} 
                  &         top-5    &  95.03    & 94.86     \\ \hline
\end{tabular}
\end{center}
\caption{Image classification results based on the original convolutional features and the poisoned convolutional features, by feeding them to the pre-trained classification decider, respectively.\label{tab_cls_eq}}
\end{table}

According to the Step 2 in Sec.~\ref{sec:step2}, we design a specific DPM ($P$) with 4 residual blocks from ResNet and insert it to the pre-trained model. It is optimized based on Eq.~(\ref{eq_optim}) with the hyper-parameter $\lambda=1.0$. 

As shown in Table~\ref{tab_cls_eq}, when we feed the original features $\varphi_1(x)$ and the poisoned features $P(\varphi_1(x))$ to the pre-trained classification decider $\varphi_2$, the achieved performance from the same architecture for image classification (top-1 and top-5 accuracy) is quite close, which indicates that the poisoned features are functionally equivalent to the original features for the specific image classification.

\begin{table}[!b]
\small
\begin{center}
\begin{tabular}{l|l|l|l|l}
\hline
\multirow{2}{*}{ $conv4\_1$} & \multicolumn{2}{c|}{L1 Distance ($\uparrow$)} & \multicolumn{2}{c}{SSIM ($\downarrow$)} \\ \cline{2-5} 
                  &  Original & Poisoned  &  Original & Poisoned  \\ \hline
ResNet50          & 0.0443    & 0.2928    & 0.6730    & 0.0070    \\ \hline
ResNet101         & 0.0406    & 0.2886    & 0.7009    & 0.0069    \\ \hline
\end{tabular}
\end{center}
\caption{Quantitative comparison of image reconstruction results from the original and poisoned features.
\label{tab_rec}}
\end{table}

\begin{figure}[!t]
    \begin{center}
    \includegraphics[width=0.95\linewidth]{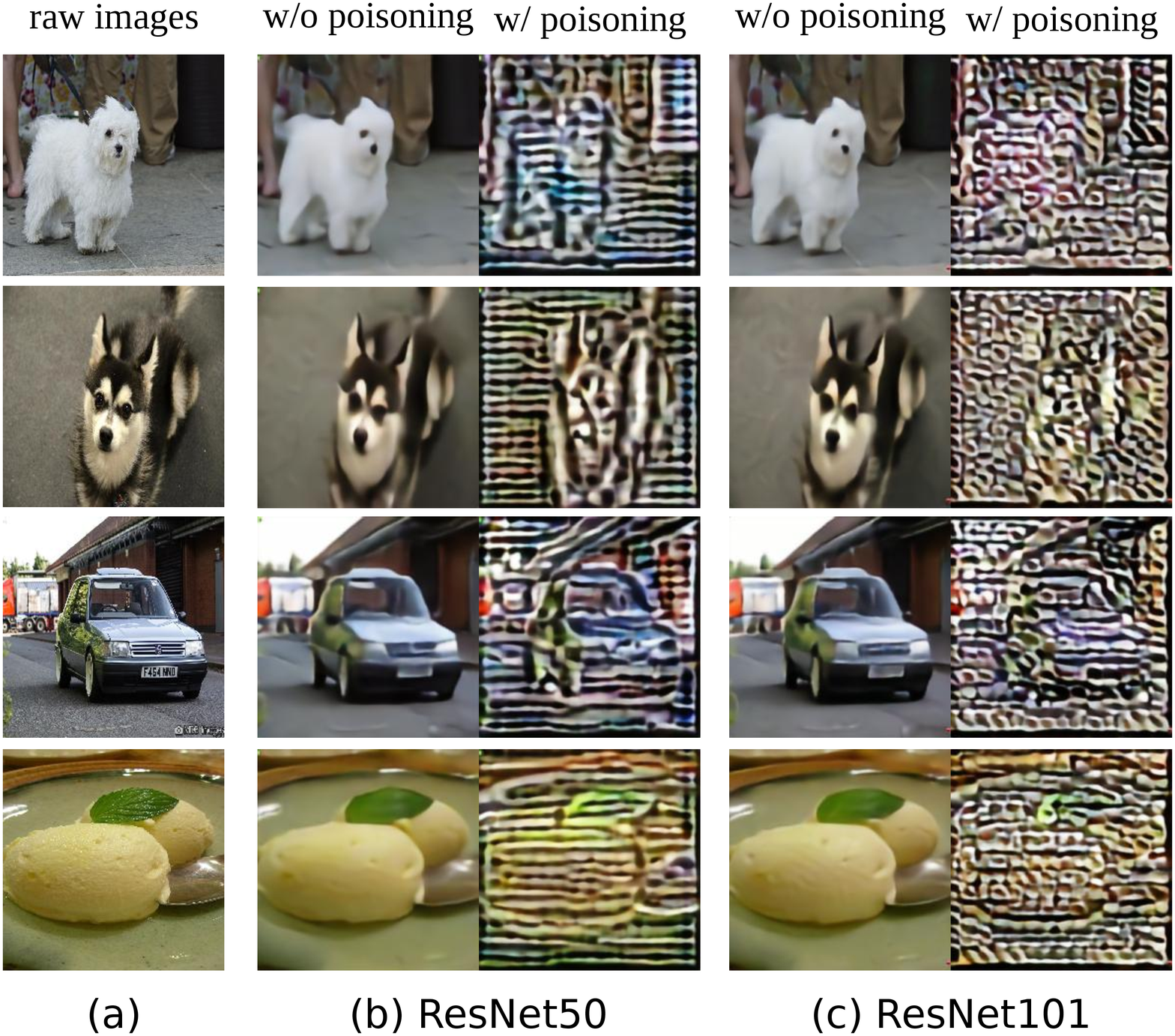}
    \end{center}
    \caption{Qualitative comparison of image reconstruction from the original convolutional features (second and fourth columns) and the poisoned convolutional features (third and fifth columns).}
    \label{fig:fig_rec_comp}
\end{figure}

Meanwhile, we use the L1 distance (Eq.~(\ref{eq_l1_loss})) and SSIM (Eq.~(\ref{eq_rec_loss})) to quantify the similarity between the reconstructed images and the original images. As shown in the second and fourth columns of Table~\ref{tab_rec}, the image reconstructor is learned as an excellent inverse to the featurizer.
It well recovers images from the original features.
As the DPM is adopted to poison the features, the inverse of the featurizer can not recover the images (third and fifth columns in Table~\ref{tab_rec}). The increasing L1 distance and decreasing SSIM indicate that the image similarity between the reconstructed images and the original images is reduced.
While the DPM is not shared to collaborators and the original images are prohibited to sharing, collaborators are prevented from learning image reconstructors to reverse the DPM. Thus, we use the pre-trained image reconstructor (i.e. the inverse of the image featurizer) to recover images from the poisoned features.
Besides, Fig.~\ref{fig:fig_rec_comp} visually compares the image reconstruction results from the original features and the poisoned features. The image contents in the reconstructed images from the poisoned features can hardly be recognized by human eyes. The visual disclosure of sensitive contents is well defended
if these images contain sensitive contents.

To summarize, the above experimental results demonstrate that the learned DPM poisons the features so that the poisoned features: 1) are classification equivalent to the original features, and 2) defend against visual disclosure of sensitive image contents when being shared.

\vspace*{-1mm}

\subsubsection{Sufficiency of DPM}

\begin{figure*}[!t]
    \begin{center}
    \includegraphics[width=\linewidth]{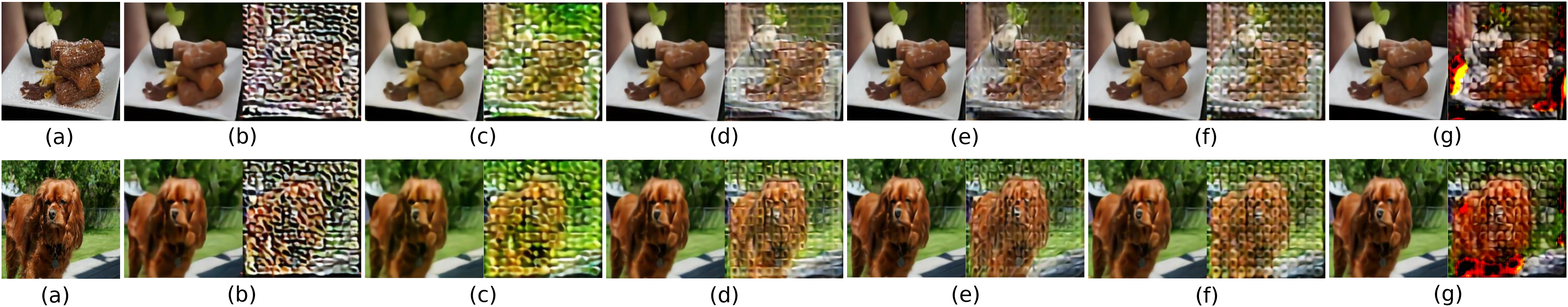}
    \end{center}
    \caption{Comparison of image reconstruction results from the original (left columns) and poisoned (right columns) convolutional features by reconstructors: (b) $px_2s_2$, (c) $px_4s_2$, (d) $rx_2s_2$, (e) $rx_4s_2$, (f) $rx_4s_4$ and (g) $rx_2s_2\_c$. (a): raw images.}
    \label{fig:fig_var_arch}
\end{figure*}

During the above DPM training and evaluation, the same fixed reconstructor is defended. This pre-trained reconstructor is an easy objective to optimize against, and in practice, there may be different networks to reverse the featurizer. Therefore, we further design and train five more reconstructors based on different architectures to reverse the pre-trained featurizer. These reconstructors are only used for testing the defensive ability of the DPM, but not used for DPM training.
We denote the only reconstructor used for DPM training as $px_2s_2$, where $p$ indicates the type of blocks used for building the reconstructor (in this case, a plain inverse bottleneck block without residual operation), $x_2$ represents two blocks before upscaling, and $s_2$ means that the upscaling factor is 2.
Similarly, other reconstructors are denoted as $px_4s_2$, $rx_2s_2$, $rx_4s_2$, $rx_4s_4$ and $rx_2s_2\_c$, where $r$ indicates inverse residual bottleneck blocks, and $c$ means the normalization strategy during reconstructor training is clamp instead of min-max normalization.
We feed the features produced by $\varphi_1$ and their corresponding poisoned features created with $P$ to each of the above reconstructors.
The reconstruction results in Fig.~\ref{fig:fig_var_arch} indicate that the learned DPM can defend various reconstructors, which have not been learned to defend during its training.

\subsubsection{Comparison with Existing Methods}
In our design, each entity learns an individual DPM to poison image features for image data sharing. To clarify the advantage of the proposed DPM, we compare it to existing approaches for addressing visual disclosure of image contents, including stationary perturbations, such as Gaussian filter (GF), Gaussian noise (GN) and mean filter (MF), and adversarial representation learning, e.g. DeepObfuscator~\cite{li2019deepobfuscator}.

\begin{figure*}[!t]
    \begin{center}
    \includegraphics[width=0.8\linewidth]{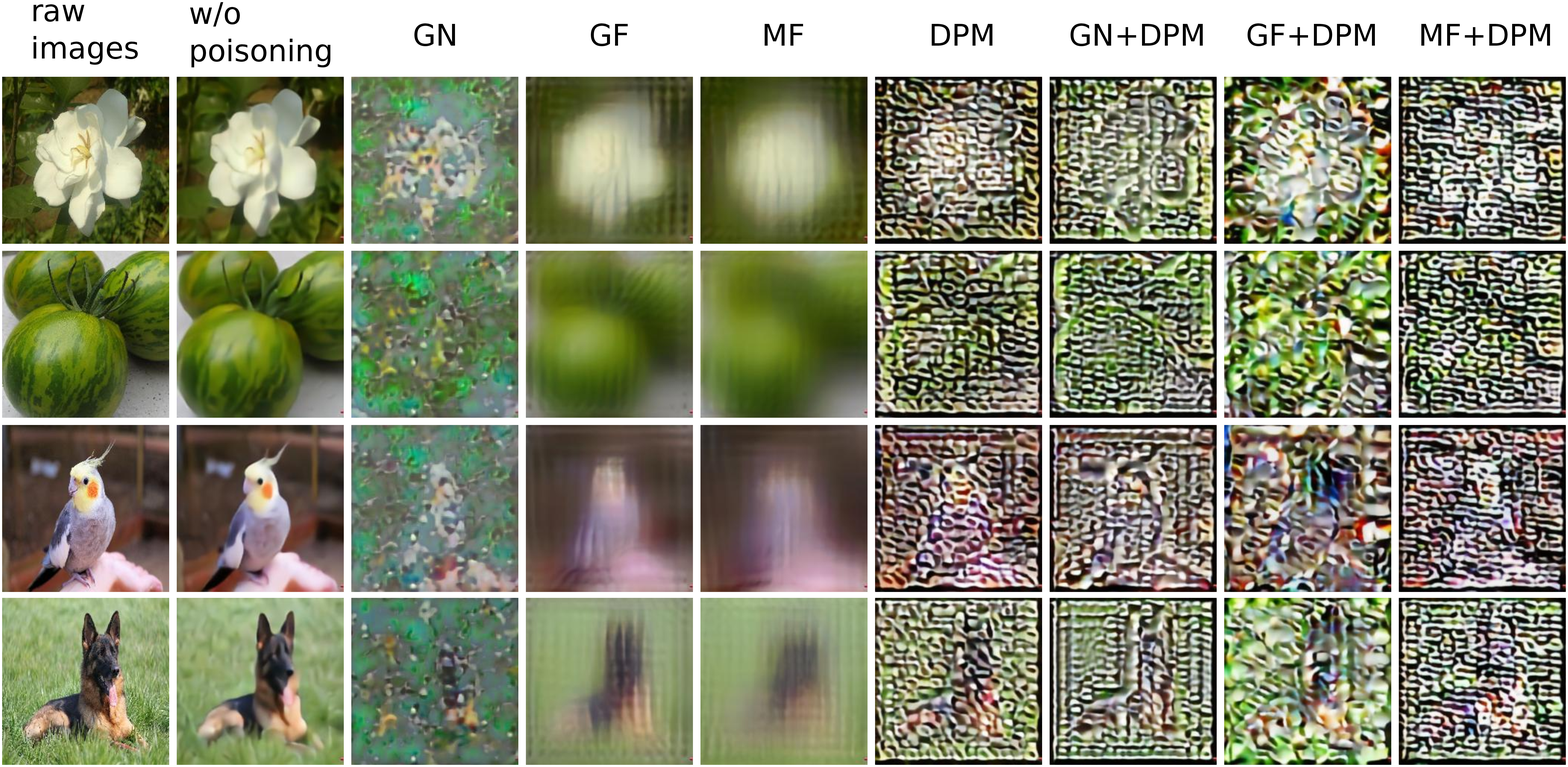}
    \end{center}
    \caption{Comparison of image reconstruction from different poisoning operations.}
    \label{fig:fig_sdpf_comp}
\end{figure*}

\begin{table}[!b]
\small
\begin{center}
\begin{tabular}{l|cc|cc}
\hline
\multirow{2}{*}{Poisoning} & \multicolumn{2}{c|}{classification (\%)} & \multicolumn{2}{c}{reconstruction} \\ \cline{2-5} 
                  & top-1     & top-5     & L1        & SSIM      \\ \hline
w/o               & 81.13     & 95.03     & 0.0406    & 0.7009    \\ \hline
GN                & 25.47     & 45.99     & 0.1905    & 0.2635    \\ 
GF                & 15.60     & 30.24     & 0.1055    & 0.4699    \\ 
MF                & 4.44     & 10.78     & 0.1169    & 0.4334    \\ \hline
DPM               & 80.78     & 94.86     & 0.2886    & 0.0069    \\ \hline
GN+DPM            & 78.10     & 93.64     & 0.3339    & 0.0047    \\ 
GF+DPM            & 78.77     & 93.78     & 0.3450    & 0.0041    \\ 
MF+DPM            & 71.23     & 89.96     & 0.3564    & 0.0186    \\ \hline
\end{tabular}
\end{center}
\caption{Result comparison between perturbations and the proposed DPM for remaining image classification performance and defending image reconstruction simultaneously.}
\label{tab_sdpf}
\end{table}

By replacing the DPM with directly applying the above stationary perturbations to convolutional features and feeding the perturbated features to the classification decider and the image reconstructor, the results shown in the top part of Table~\ref{tab_sdpf} indicate that the perturbations can defend against image reconstruction to some extent, but also suppress the ability of the features for image classification. While the proposed DPM can achieve classification equivalence between poisoned features and original features, it achieves more promising results for both remaining image classification performance and defending against image reconstruction. Furthermore, combining stationary perturbations with DMP still achieves promising results -- defending against the image reconstruction with minimal loss in classification performance, as shown in the bottom rows of Table~\ref{tab_sdpf}. Meanwhile, the visual comparison of image reconstruction from image features altered by stationary perturbations and the proposed DPM in Fig.~\ref{fig:fig_sdpf_comp} further validates the advantage of the proposed DPM over conventional perturbations.

\begin{table*}[t]
\small
\centering
\begin{tabular}{l|c|c|cc||cc}
\hline
 & utility recognition  & privacy recognition  & \multicolumn{2}{c||}{privacy reconstruction} & \multicolumn{2}{c}{image sharing reconstruction} \\ 
 & mAPs ($\uparrow$, \%) & mAPs ($\downarrow$, \%) &    L1 ($\uparrow$) & SSIM ($\downarrow$)  & L1 ($\uparrow$)  &    SSIM ($\downarrow$)  \\ \hline
w/o protection & 81.08 & 80.54 &   0.0200        & 0.9177          &    0.0200        & 0.9177    \\ \hline
DeepObfuscator~\cite{li2019deepobfuscator} & 73.71 & 65.07 &    0.2614       & 0.3997          &    0.0359       & 0.8232          \\ \hline
DPM & 79.53 & 65.95 &   0.2573        & 0.0064          &    \multicolumn{2}{c}{Denied}   \\ \hline
\end{tabular}
\caption{Result comparison between the privacy-preserving obfuscation and the proposed DPM.}
\label{tab_obfuscation}
\end{table*}

\begin{figure}
    \centering
    \includegraphics[width=0.9\linewidth]{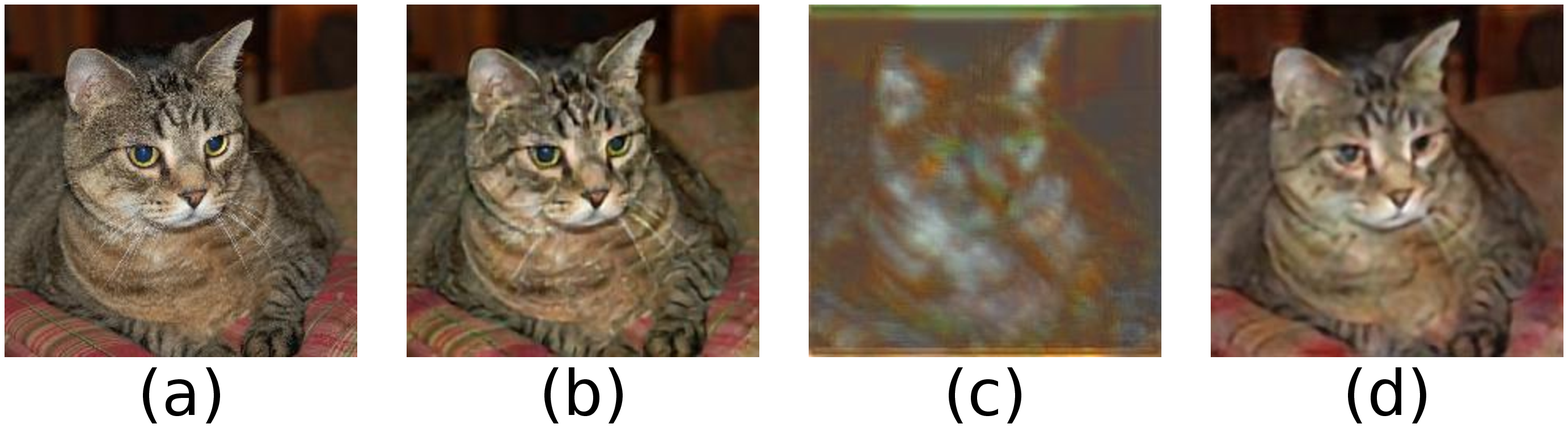}
    \caption{(a) the original image; (b) image reconstruction from the non-poisoned features; (c) DeepObfuscator defending its adversarial reconstructor; (d) image reconstruction from obfuscated features (DeepObfuscator) by reversing the obfuscator.}
    \label{fig:fig_share_rec}
\end{figure}

Additionally, we compare the proposed method with a privacy-preserving method, i.e. DeepObfuscator~\cite{li2019deepobfuscator} for image reconstruction under both conventional privacy preserving and the introduced image data sharing for collaboration. The comparison experiment is conducted on CelebA dataset~\cite{liu2015faceattributes} for face attribute recognition. Similar to~\cite{li2019deepobfuscator}, the first 20 attributes are regarded as the utility in privacy preserving or the vision task to be addressed in image data sharing, while the remaining 20 attributes are the privacy, and the image reconstruction is another privacy and visual disclosure for image data sharing. The VGGNet~\cite{simonyan2014very} is adopted as the backbone for fair comparison and we split the model at the same point as DeepObfuscator. As shown in columns 2 to 5 of Table~\ref{tab_obfuscation}, the proposed DPM achieves better overall results on privacy preserving. Specifically, the DPM better remains the utility recognition and defends the reconstruction that leaks privacy.
When being adopted to image data sharing for task-specific collaboration against visual disclosure, sharing the obfuscator of~\cite{li2019deepobfuscator} among collaborators results in the reversing of the obfuscator, which makes the obfuscated features reconstructed to the original images, as shown in Fig.~\ref{fig:fig_share_rec}. Also, in columns six and seven of Table~\ref{tab:tab_collaboration}, when an entity learns an image reconstructor to recover images from the original image features, the reconstructed images are highly similar to the original images, with L1 distance of 0.0200 and SSIM of 0.9177. Similarly, reversing the obfuscator of DeepObfuscator also achieves great reconstruction quality, which is not desired. This is because there is a large overlap between the recognition-related information and reconstruction-relation information in the convolutional features. To guarantee the recognition performance, the reconstruction-related informaiton can not be removed thoroughly from the image features.
Compared with DeepObfuscator, the proposed DPM allows each entity to keep its own DPM in private and maintain the poisoned features from different DMPs in the same feature space. Keeping DPM in private denies reversing it for the image reconstruction from poisoned features.
Thus, the proposed DPM can also be used for privacy preserving, while the privacy-preserving methods can not address the introduced task of image data sharing against visual disclosure.

\subsection{Entity Collaboration -- Model Refinement}
%

In this section, we conduct experiments to verify that the poisoned features shared from each entity helps others improve the image classification performance.
To simulate the collaborators with limited training images, we define three subsets of images from the image set $\mathbb{S}$, each of which is assumed to be collected by an entity (in $\{E_1, E_2, E_3\}$) with 10\% images (64056 images) from $\mathbb{S}$, denoted as $\mathbb{S}_1$, $\mathbb{S}_2$ and $\mathbb{S}_3$. Besides, a fourth entity ($E_4$) collects 1\% images of $\mathbb{S}$, denoted as $\mathbb{S}_4$. As shown in rows (1 -- 4) of Table~\ref{tab:tab_collaboration}, when various entities learn classification models individually from their own images, respectively, the classification performance is severely limited by the number of training images.

Suppose the first entity uses $\mathbb{S}_1$ pre-trains the classification network and splits it according to Sec.\ref{sec:step1}, then each of the entity follows the similar procedure to train individual DPMs ($P_1$, $P_2$, $P_3$ and $P_4$) for feature poisoning as Sec.~\ref{sec:step2}. A shared pool of image features among various collaborators is established as: 
$\{
P_1(\varphi_1(\mathbb{S}_1)),
P_2(\varphi_1(\mathbb{S}_2)),
P_3(\varphi_1(\mathbb{S}_3)),
P_4(\varphi_1(\mathbb{S}_4))
\}$.
Then, each entity can combine its own image data and the image data shared from other entities to refine the classification decider $\varphi_2$, as shown in Fig.~\ref{fig:fig_intro}(b). 

As shown in the rows 7 and 8 in Table~\ref{tab:tab_collaboration}, two entities $E_1$ and $E_2$ use the poisoned image features from each other and achieve much better classification performance than training models based on its own data (rows 1 and 2), respectively. Compared with combining image features directly (row 6), there is certain loss in performance improvement due to the poisoning operation leading to information loss. Besides, when the number of collaborators increases, e.g. $E_3$ (in rows 9 and 10) exploits data from $E_1$ and $E_2$, the performance can be further improved and is much better than just using its own data $\mathbb{S}_3$ in row 3. Specially, when entity $E_4$ train its classification model based on its own images, the top-1 accuracy is 13.5\%. With other entities sharing image data to $E_4$, it can easily achieve 61.6\% top-1 accuracy (in row 12). These comparison results indicate that the image data sharing does benefit entities since they can use others' data for its model refinement.

\begin{table}[!t]
\small
\begin{center}
\begin{tabular}{l|p{4cm}|c|c}
\hline
rows & image data & top-1 & top-5 \\ \hline
1 -- $E_1$& raw images $\mathbb{S}_{1}$ (10\%) & 52.9 & 77.7 \\ 
2 -- $E_2$& raw images $\mathbb{S}_{2}$ (10\%)& 52.7 & 77.8 \\ 
3 -- $E_3$& raw images $\mathbb{S}_{3}$ (10\%)& 53.2 & 77.9 \\ 
4 -- $E_4$& raw images $\mathbb{S}_{4}$ (1\%) & 13.5 & 30.3 \\ 
5& raw images $\mathbb{S}$ (100\%) & 81.1 & 95.0 \\ \hline \hline 
6& $\varphi_1(\mathbb{S}_{1}) \cup \varphi_1(\mathbb{S}_{2})$ & 61.2 & 83.8 \\  \hline
7 -- $E_1$&  $\varphi_1(\mathbb{S}_{1}) \cup P_2(\varphi_1(\mathbb{S}_{2}))$ & 59.4 & 82.6 \\
8 -- $E_2$&  $P_1(\varphi_1(\mathbb{S}_{1})) \cup \varphi_1(\mathbb{S}_{2})$ & 59.8 & 83.0 \\ \hline \hline
9 -- $E_3$&  $P_1(\varphi_1(\mathbb{S}_{1})) \cup P_2(\varphi_1(\mathbb{S}_{2})) \cup \varphi_1(\mathbb{S}_3)$ & 62.3 & 84.5 \\
10 -- $E_3$&  $P_1(\varphi_1(\mathbb{S}_{1})) \cup P_2(\varphi_1(\mathbb{S}_{2})) \cup P_3(\varphi_1(\mathbb{S}_3)) \cup \varphi_1(\mathbb{S}_3)$ & 62.7 & 84.7 \\ \hline \hline
11 -- $E_4$&  $P_1(\varphi_1(\mathbb{S}_{1})) \cup P_2(\varphi_1(\mathbb{S}_{2})) \cup \varphi_1(\mathbb{S}_4)$ & 58.4 & 82.1 \\ 
12 -- $E_4$&  $P_1(\varphi_1(\mathbb{S}_{1})) \cup P_2(\varphi_1(\mathbb{S}_{2})) \cup P_3(\varphi_1(\mathbb{S}_3)) \cup \varphi_1(\mathbb{S}_4) $ & 61.6 & 84.4 \\ \hline \hline
\end{tabular}
\end{center}
\caption{
Classification performance of image classification models trained/fine-tuned on image data in different representations.
\label{tab:tab_collaboration}
}
\end{table}

\section{Conclusion and Future Work}

This paper introduced a new vision task of image data sharing against visual disclosure of sensitive contents. The task aims to make various entities refine respective models by using image data shared by others, but not visually observe the sensitive content in the shared data. 
To achieve this goal, we proposed that each entity inserts an extra Deep Poisoning Module to the pre-trained network. The DPMs were learned to make the poisoned features be functionally equivalent to the non-poisoned features and defend against the image reconstruction. 
Being kept in private, DPMs provided a structure-based approach to prevent the shared image data from visually disclosing sensitive image contents.
Our experiments verified the effectiveness of the proposed method by simulating the process of sharing data to benefit collaborators' model refinement without visually disclosing sensitive image contents.

Besides, this paper is an initial exploration of the introduced task, there are still many problems to be addressed, such as the potential non-i.i.d. issue among images from various entities, expanding to other vision applications, etc. These issues may be discussed in the future work.

\newpage 
{\small
\bibliographystyle{ieee_fullname}
\bibliography{dpf_ref}
}

\end{document}


\title{Deep Poisoning: Towards Robust Image Data Sharing against Visual Disclosure}

\author{First Author\\
Institution1\\
Institution1 address\\
{\tt\small firstauthor@i1.org}
\and
Second Author\\
Institution2\\
First line of institution2 address\\
{\tt\small secondauthor@i2.org}
}

\maketitle

\subsection*{ImageNet Split}
As mentioned in Sec. 4 of the paper, we split the ImageNet~\cite{ILSVRC15} into the private and public sets. We will release the category split files later along with the source code of this paper.

\subsection*{More Reconstruction Results}

\begin{figure}[!t]
    \centering
    \includegraphics[width=0.8\linewidth]{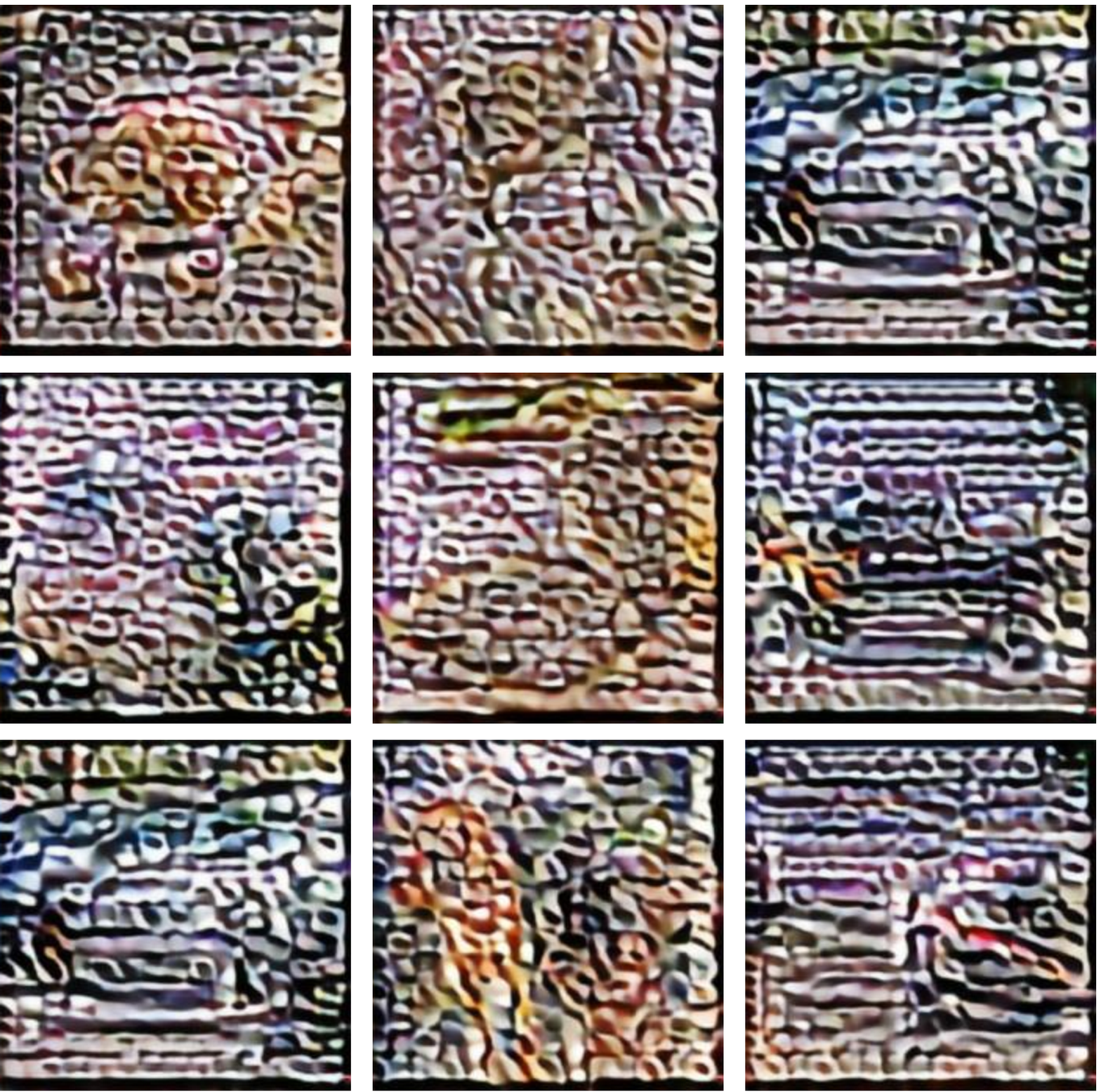}
    \caption{Illustrations of reconstruction results from the reconstructor, $px_2s_2$.}
    \label{fig:fig_supp_px2s2}
\end{figure}

\begin{figure}[!t]
    \centering
    \includegraphics[width=0.8\linewidth]{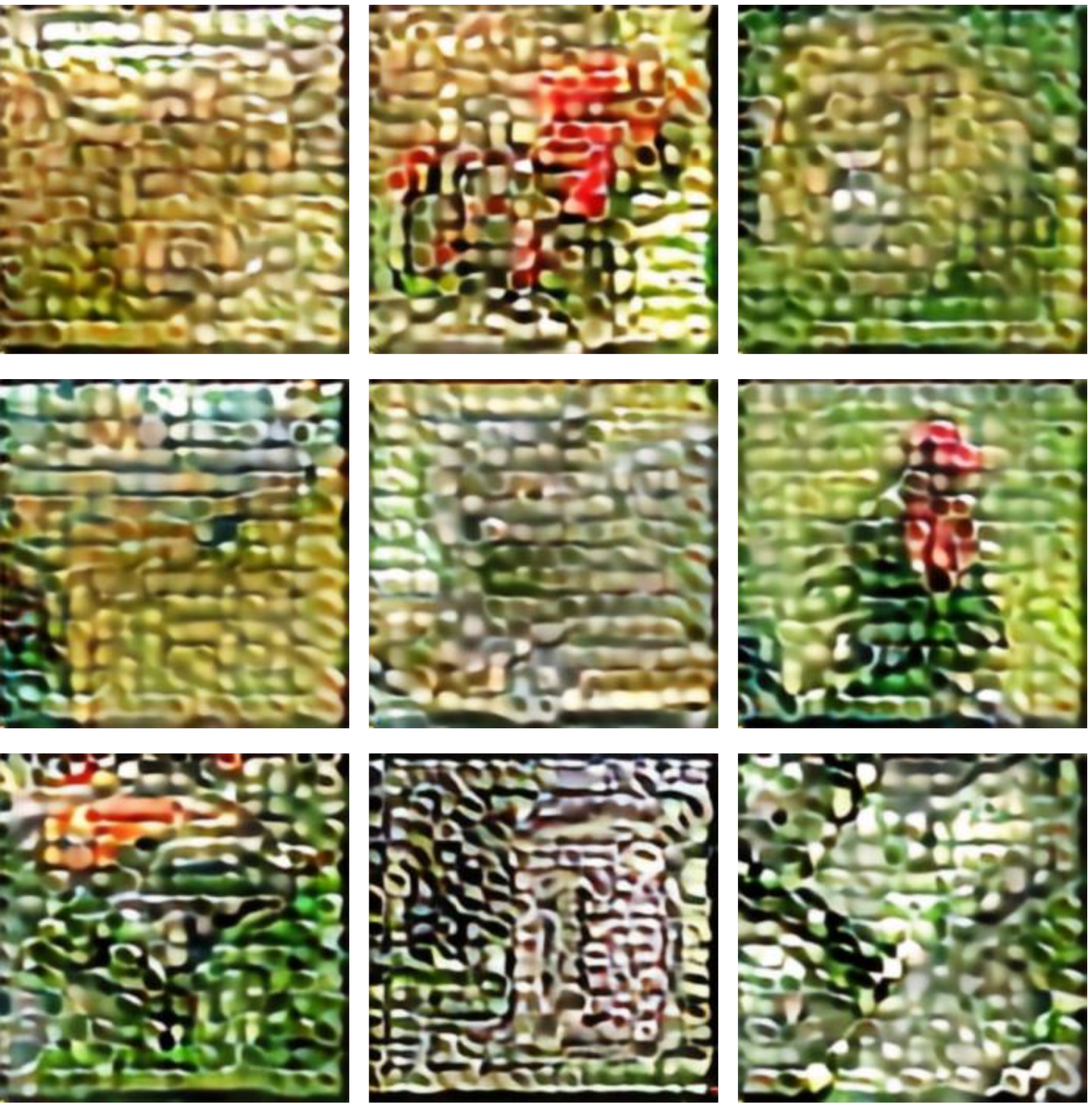}
    \caption{Illustrations of reconstruction results from an unknown reconstructor, $px_4s_2$.}
    \label{fig:fig_supp_px4s2}
\end{figure}

\begin{figure}[!t]
    \centering
    \includegraphics[width=0.8\linewidth]{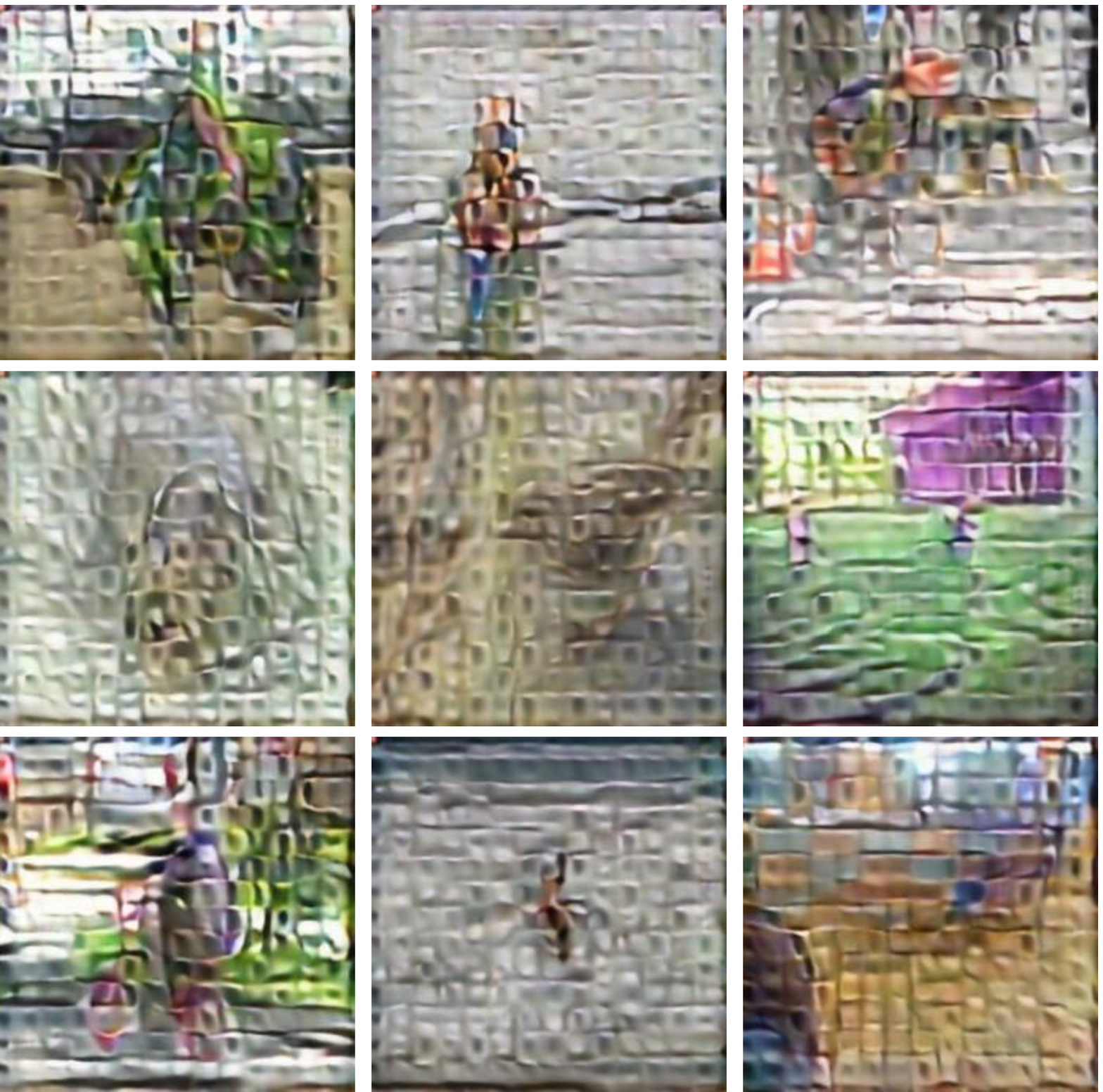}
    \caption{Illustrations of reconstruction results from an unknown reconstructor, $rx_2s_2$.}
    \label{fig:fig_supp_rx2s2}
\end{figure}
We compare the raw images, reconstructions from the non-poisoned and poisoned convolutional features in the paper. After observing the raw images, people may guess the content in the reconstructed images according to the instant memory of the raw images. Therefore, we also present some reconstructed images from the poisoned convolutional features without providing the original images. The DPF is learned to poison the convolutional features inferred by $conv4\_1$ of ResNet101. Several reconstruction results from the initial reconstructor ($px_2s_2$) are shown in Fig.~\ref{fig:fig_supp_px2s2}, while some other reconstructions from $px_4s_2$ and $rx_2s_2$, which are unknown during the DPF training, are shown in Fig.~\ref{fig:fig_supp_px4s2} and Fig.~\ref{fig:fig_supp_rx2s2}, respectively.
These reconstruction results further verify that the DPF suppresses the reconstruction-related information in the  convolutional features. The visual perception of the contents in the reconstructed images is substantially impaired.

\subsection*{Architectures}

\begin{figure}[!t]
    \centering
        \centering
        \includegraphics[width=0.7\linewidth]{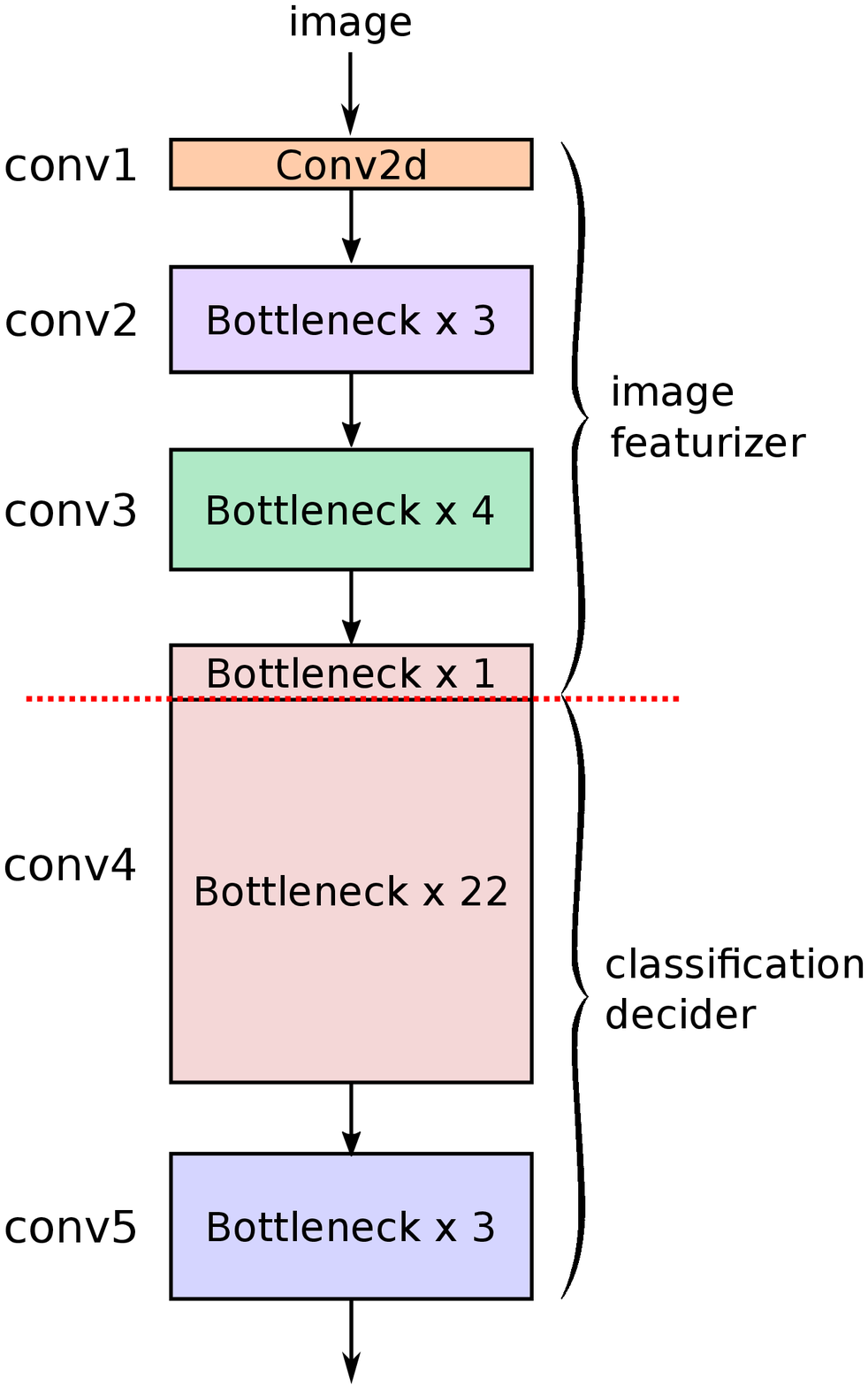}
        \caption{Image featurizer and classification decider split from ResNet101 at hook point $conv4\_1$.}
        \label{fig:supp_arch}
\end{figure}

\begin{figure}[!t]
        \centering
        \includegraphics[width=0.85\linewidth]{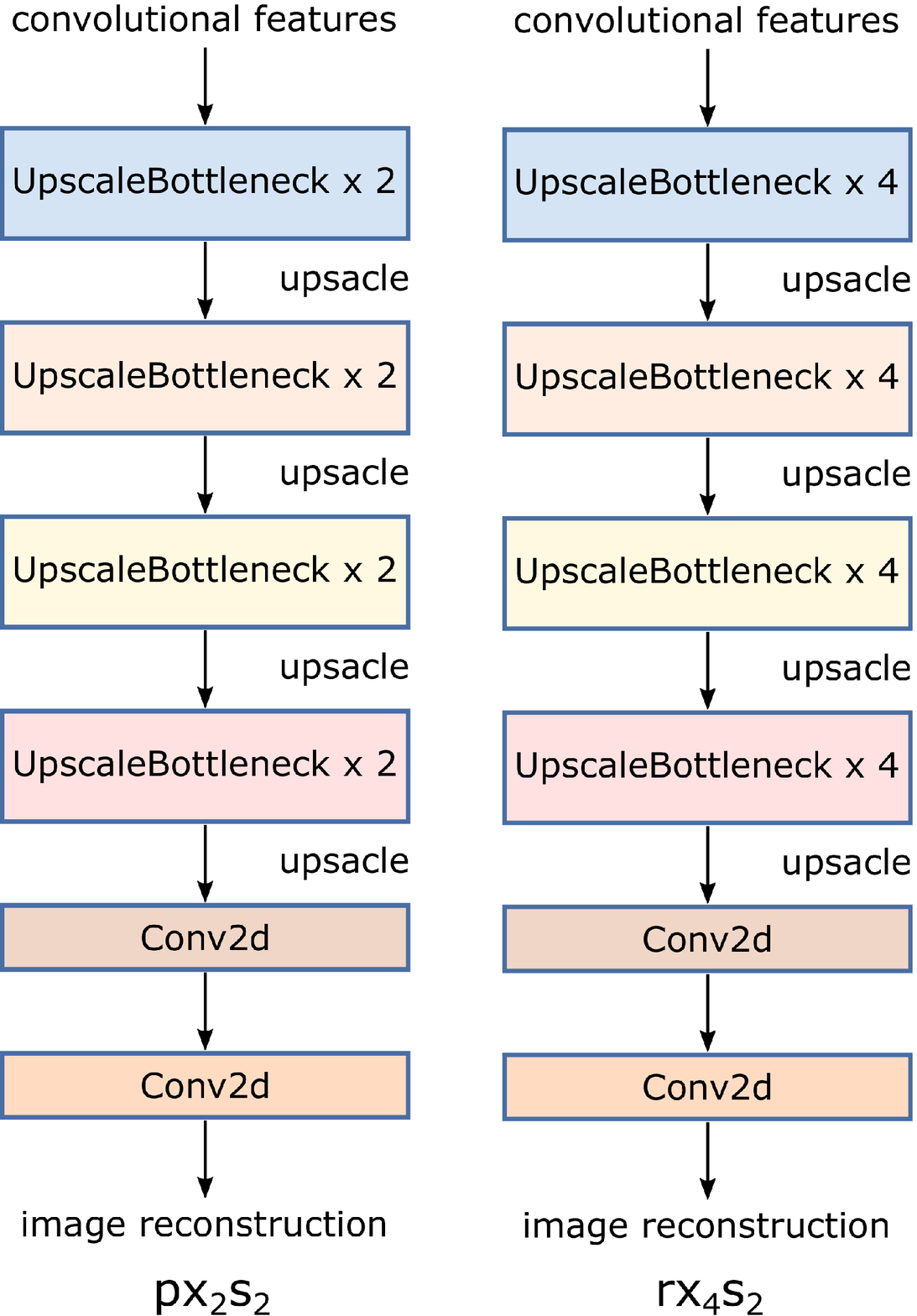}
        \caption{Illustrations of reconstructors for the same convolutional features}
        \label{fig:supp_rec_arch}
\end{figure}

\begin{figure*}[!t]
    \centering
    \includegraphics[width=0.8\linewidth]{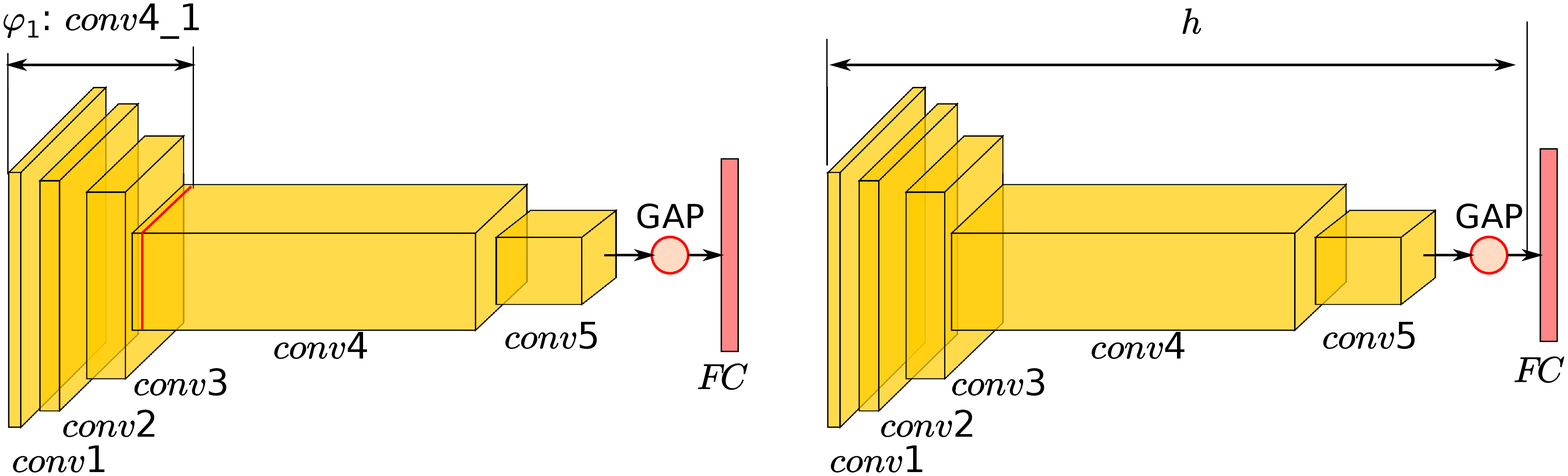}
    \caption{Different featurizer depth splits on the ResNet architecture.}
    \label{fig_feat_split}
\end{figure*}

\begin{table*}[!t]
\caption{Classifier refinement based on different training image data.\label{tab_refinement}}
\centering
\begin{tabular}{l|ccc||cc||c}
\hline
training data & $\mathbb{S}_1$ & $\mathbb{S}_2$ & $\mathbb{S}_1\cup\mathbb{S}_2$ & $h(\mathbb{S}_1\cup\mathbb{S}_2)$ & $\varphi_1(\mathbb{S}_1\cup\mathbb{S}_2)$ & $P(\varphi_1(\mathbb{S}_1)) \cup \varphi_1(\mathbb{S}_2)$ \\ \hline
top-1 (\%) & 52.9 & 52.7 & 58.8 & 54.7 & 61.2 & 59.8 \\ \hline
top-5 (\%) & 77.9 & 77.8 & 82.1 & 79.2 & 83.8 & 83.0 \\ \hline
\end{tabular}
\end{table*}

During our experiments, we mainly use the ResNet~\cite{he2016deep} architecture as our backbone to train the initial image classification models. We use the hook point $conv4\_1$ to split the pre-trained model into image featurizer and classification decider.
For example, when the ResNet101 is used as the backbone and the hook point is set to $conv4\_1$, the architecture is split as shown in Fig.~\ref{fig:supp_arch}. To reconstruct the images from convolutional features, the initial reconstructor $px_2s_2$ and an additional reconstructor $rx_4s_2$ are shown in Fig.~\ref{fig:supp_rec_arch}. The ``UpscaleBottleneck" is an inverse to the Bottleneck module in constructing the traditional ResNet. The residual operations are configurable during creating the upscale modules.

Besides, to achieve image data sharing without visual disclosing sensitive contents, another straightforward way is to increase the depth of the image featurizer. As the depth is increased, the spatial dimension of the output features may be decreased to a level, where the spatial information is drastically suppressed for image reconstruction. For example, as shown in Fig.~\ref{fig_feat_split}, instead of using the layers up to $conv4\_1$ as the featurizer $\varphi_1$, one can also use all layers before the final Fully Connected layer (FC, the linear layer in the ResNet architecture) as the featurizer $h$. $h$ outputs features in the spatial dimension of $1 \times 1$, which is hard to be used for image reconstruction.

However, to ensure the features shared from different entities to be the same task-specific, the featurizer is fixed and shared among entities. With the most layers of the model are fixed, enlarging the training data volume for model refinement (fine-tuning the classification decider), which is the target of the introduced task of image data sharing, would be impaired.
Comparing columns 5 and 6 in Table~\ref{tab_refinement} shows that sharing features from a deeper featurizer ($h(\mathbb{S}_1\cup\mathbb{S}_2)$, top-1 accuracy of 54.7\%) is much less effective than sharing features from a shallower featurizer ($\varphi_1(\mathbb{S}_1\cup\mathbb{S}_2)$, top-1 accuracy of 61.2\%) for enlarging training data volume among entities by image data sharing. Furthermore, while poisoned features defend the visual disclosure of sensitive contents, sharing poisoned features from a shallower featurizer ($P(\varphi_1(\mathbb{S}_1)) \cup \varphi_1(\mathbb{S}_2)$, top-1 accuracy of 59.8\%) achieves obvious effect on enlarging training data volume -- much better than sharing features from a deeper featurizer. A featurizer with relatively shallower depth, such as $conv4\_1$, is necessary for the proposed framework.

\subsection*{More Details on Comparison with DeepObfuscator}

\begin{figure}[!t]
    \centering
    \includegraphics[width=\linewidth]{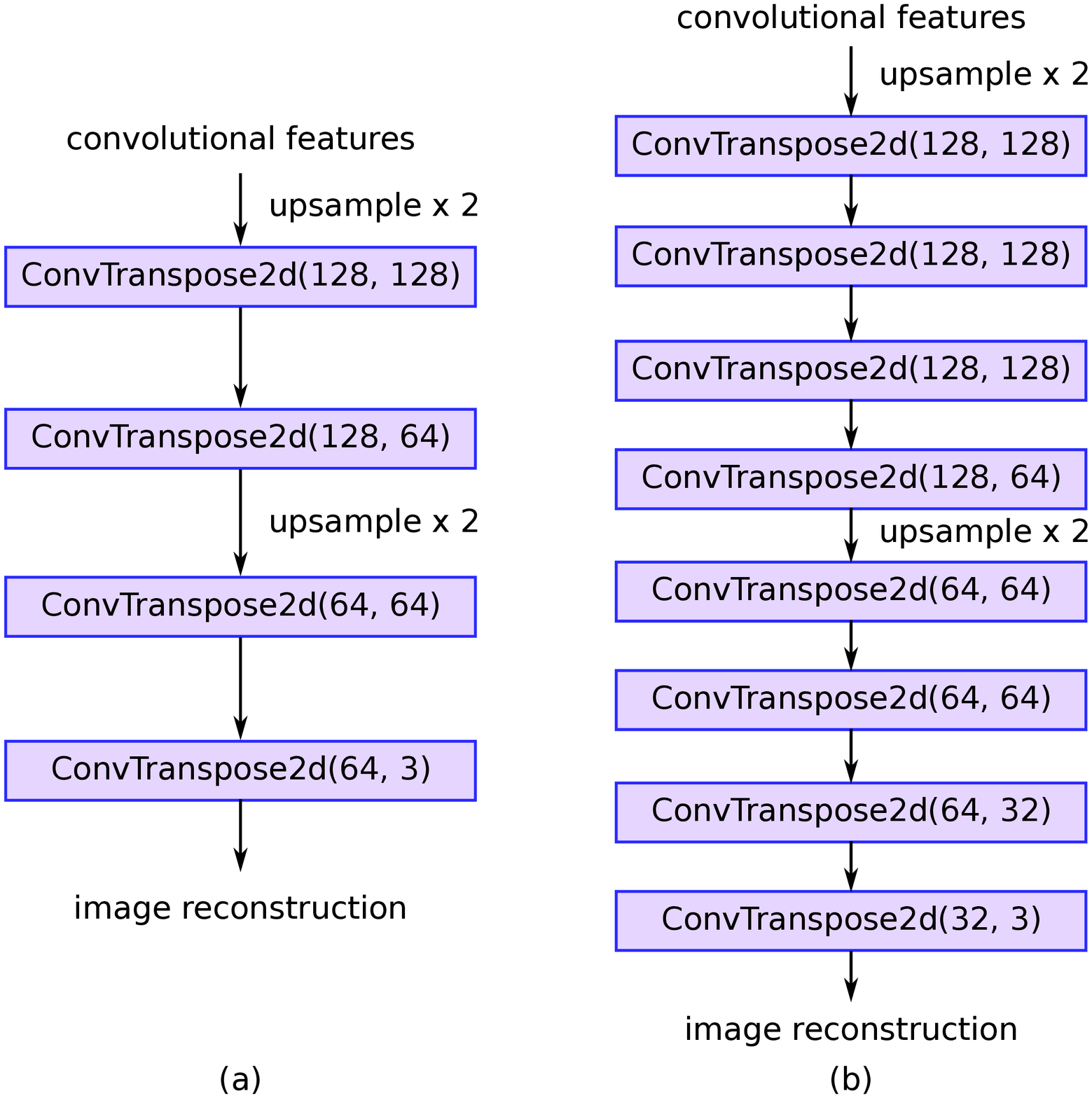}
    \caption{(a) Adversarial reconstructor from DeepObfuscator~\cite{li2019deepobfuscator}; (b) Re-trained reconstructor to reverse the obfuscated convolutional features.}
    \label{fig:fig_supp_do_archs}
\end{figure}

In Table~4 of the paper, we reproduce the DeepObfuscator~\cite{li2019deepobfuscator} for comparison.
Of the 40 attributes defined in CelebA~\cite{liu2015faceattributes}, we use the first twenty-attribute recognition as the privacy-preserving utility and the vision task to be addressed in the introduced task.
The remaining twenty-attribute recognition and image reconstruction are privacy to protect. Meanwhile, the image reconstruction from convolutional features is also the visual disclosure to be prevented in the proposed task.
We use VGG16~\cite{simonyan2014very} as the backbone, the same with the comparison method~\cite{li2019deepobfuscator}.
After pre-training the deep network for recognizing the first 20 attributes and the adversarial models (attribute recognizer and image reconstructor) for attribute leakage and image reconstruction, we follow the same training algorithm as~\cite{li2019deepobfuscator} to train the obfuscator.

Even though the deep obfuscator can defend the image reconstruction, rich visual information is still preserved by the obfuscated convolutional features. If the obfuscator is shared among entities, one entity can still re-train image reconstructor to reverse the obfuscation operation. Thus, obfuscated features shared from an entity could be reconstructed (by another entity) to the original images, which leads to the undesired visual disclosure of sensitive image contents.
The different adversarial reconstructor and the re-trained reconstructor architectures are shown in Fig.~\ref{fig:fig_supp_do_archs}.
The source code for this reproduction will also be released later along with the source code of this paper.

{\small
\bibliographystyle{ieee_fullname}
\bibliography{dpf_ref}
}